\documentclass[a4paper,twoside]{article}

\usepackage{epsfig}
\usepackage{subcaption}
\usepackage{calc}
\usepackage{amssymb}
\usepackage{amstext}
\usepackage{amsmath}
\usepackage{amsthm}
\usepackage{multicol}
\usepackage{pslatex}
\usepackage{apalike}
\usepackage{tabularx}
\usepackage[table]{xcolor}
\usepackage{algorithm2e}
\usepackage{multirow}
\usepackage{booktabs} 
\usepackage[bottom]{footmisc}
\usepackage{SCITEPRESS}     

\begin{document}

\title{Model Predictive Control for Crowd Navigation via Learning-Based Trajectory Prediction}

\author{\authorname{
Mohamed Parvez Aslam\sup{1},
Bojan Derajic\sup{2, 3}\orcidAuthor{0009-0000-6175-535X},
Mohamed-Khalil Bouzidi\sup{2, 4}\orcidAuthor{0009-0009-9734-3133}, \\
Sebastian Bernhard\sup{2}\orcidAuthor{0000-0002-7194-7539}
and Jan Oliver Ringert\sup{1}\orcidAuthor{0000-0002-3610-3920}}
\affiliation{\sup{1}Bauhaus-Universität Weimar, Germany}
\affiliation{\sup{2}Continental Automotive GmbH, Germany}
\affiliation{\sup{3}Technical University of Berlin, Germany}
\affiliation{\sup{4}Free University of Berlin, Germany}
\email{Contact e-mail: mohamed.parvez.aslam@uni-weimar.de}
}


\abstract{Safe navigation in pedestrian-rich environments remains a key challenge for autonomous robots. This work evaluates the integration of a deep learning-based Social-Implicit (SI) pedestrian trajectory predictor within a Model Predictive Control (MPC) framework on the physical \textit{Continental Corriere} robot. Tested across varied pedestrian densities, the SI-MPC system is compared to a traditional Constant Velocity (CV) model in both open-loop prediction and closed-loop navigation. Results show that SI improves trajectory prediction—reducing errors by up to 76\% in low-density settings—and enhances safety and motion smoothness in crowded scenes. Moreover, real-world deployment reveals discrepancies between open-loop metrics and closed-loop performance, as the SI model yields broader, more cautious predictions. These findings emphasize the importance of system-level evaluation and highlight the SI-MPC framework’s promise for safer, more adaptive navigation in dynamic, human-populated environments.}

\onecolumn \maketitle \normalsize \setcounter{footnote}{0} \vfill

\section{\uppercase{Introduction}}
\label{sec:introduction}

The rapid advancement of autonomous systems has transformed mobile robotics, particularly in shared environments where robots and pedestrians coexist \cite{DL_vs_KB_korbmacher_review_2022,review_mixed_env_golchoubian_pedestrian_2023}. A critical challenge in this domain is ensuring safe navigation in dynamic settings, such as crowded sidewalks and factory floors \cite{wabersich_data-driven_2023}. Pedestrian trajectory prediction (PTP) is central to this challenge, enabling robots to anticipate human movements and adjust their behavior proactively \cite{alahi_social_2016,salzmann_trajectron_2021}.

\begin{figure}[!htb]
    \centering
    \includegraphics[width=1.0\linewidth]{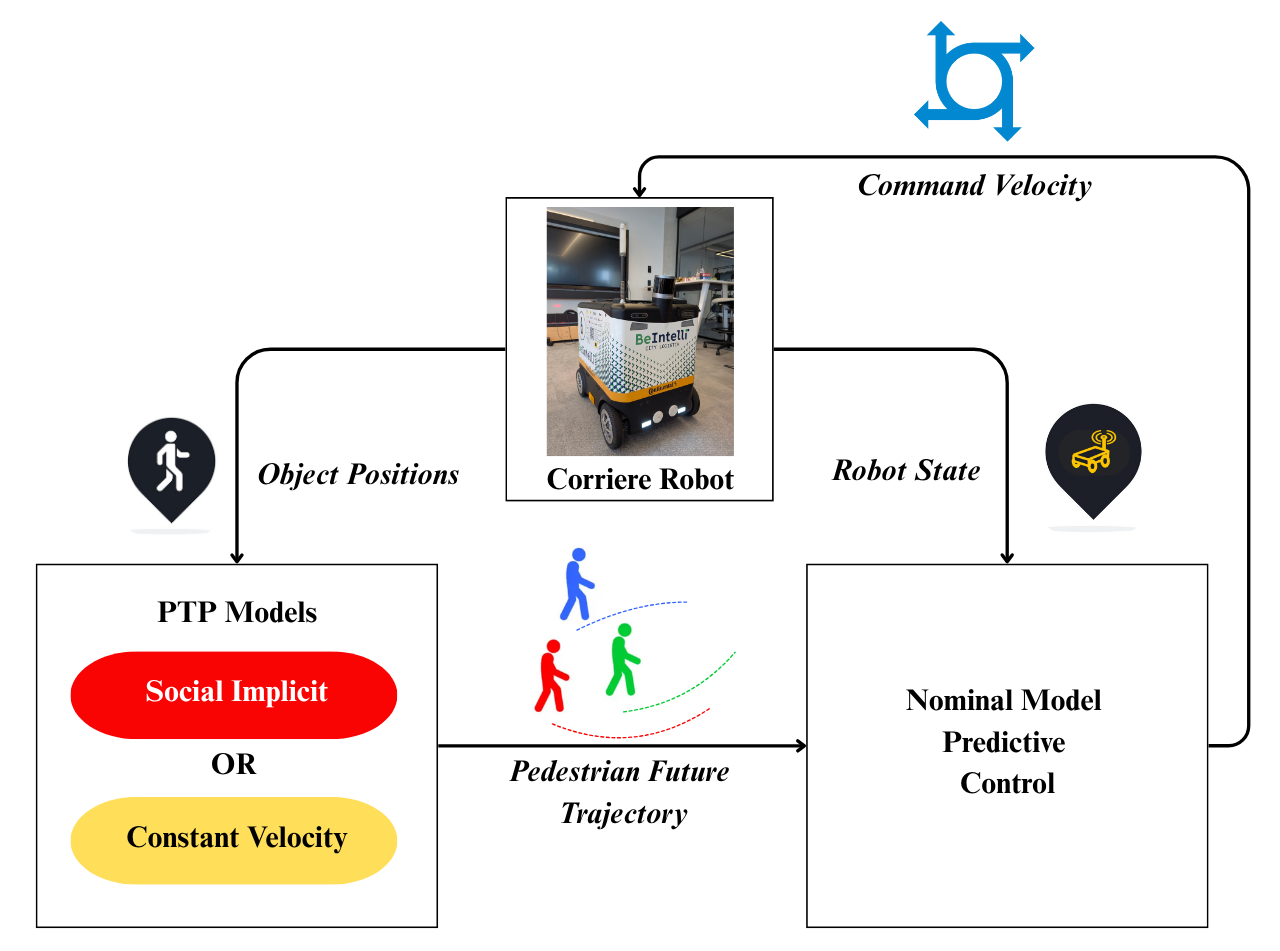}
    \caption{The architecture of an MPC-based motion planning framework with PTP module.}
    \label{fig:framework}
\end{figure}

Traditional heuristic- or knowledge-based models, while interpretable, often fail to capture the complexity of pedestrian behavior \cite{review_DL_sighencea_review_2021}. In contrast, deep learning methods—such as Social LSTM \cite{alahi_social_2016}, Social GAN \cite{huang_sti-gan_2021}, spatiotemporal graph models \cite{salzmann_trajectron_2021,mohamed_social-stgcnn_2020,lian_ptp-stgcn_2023}, and hybrid architectures like Y-Net \cite{mangalam_goals_2020} and Social-Implicit (SI) \cite{mohamed_social-implicit_2022}—leverage datasets like ETH/UCY \cite{5459260} to model social interactions, generally outperforming traditional approaches in open-loop accuracy.

However, open-loop evaluation on static datasets overlooks how these models perform in real-world, dynamic environments, where real-time execution and distribution shift are critical \cite{alahi_social_2016}. Despite progress in learning-based trajectory prediction, integration into full navigation stacks and closed-loop evaluation remains limited.

Model Predictive Control (MPC) provides a principled framework for dynamic navigation, relying heavily on accurate motion predictions \cite{bhatt_mpc-pf_2023}. While MPC aligns well with the output of PTP models, many systems still use simplistic predictors like constant velocity \cite{degroot2023globallyguidedtrajectoryplanning,brito,constvel1} or other heuristic-based methods \cite{chen,piccinelli_mpc_2023,mukhtar}.

Our contributions in this paper are summarized below:
\begin{itemize}
    \item First, we identify the criteria required for a learning-based PTP model to be integrated within an MPC framework and deploy it on a physical robot, as shown in Fig.~\ref{fig:framework}. The resulting navigation behavior is evaluated by quantifying improvements in both safety metrics---such as minimum collision distance---and trajectory smoothness, measured via acceleration rate~\cite{francis2023principlesguidelinesevaluatingsocial}.
    \item Second, we compare the open-loop performance of predictors on standard datasets with their behavior when integrated into a full navigation stack, thereby assessing the generalization capacity of learned models.
    \item Third, we demonstrate through extensive experiments that learning-based predictors provide tangible improvements over the baseline, leading to smoother trajectories and improved safety. The evaluation assessment is presented, measuring prediction accuracy, spread, smoothness, and safety across varying pedestrian densities.
\end{itemize}

\section{Preliminaries}

\subsection{Problem Formulation}

We consider a mobile robot navigating in a dynamic, pedestrian-populated environment. The goal is to generate a collision-free, dynamically feasible trajectory that minimizes a cost function over a finite horizon. The optimization incorporates the robot’s constraints and learning-based pedestrian motion predictions that model human interactions. The robot adapts its trajectory in real time to ensure safe and socially compliant navigation.

\subsection{Trajectory Prediction Metrics}

To evaluate predicted trajectories, we use four key metrics:

Average Displacement Error (ADE) measures the mean Euclidean distance between predicted and ground truth positions over the prediction horizon:
\begin{equation}
    \text{ADE} = \frac{1}{N} \sum_{t=1}^{N} \| \hat{p}_t - p_t \|
\end{equation}
where \( \hat{p}_t \) and \( p_t \) are the predicted and true positions at time \( t \), and \( N \) is the number of future steps.
Final Displacement Error (FDE) measures the distance at the final prediction step:
\begin{equation}
    \text{FDE} = \| \hat{p}_N - p_N \|
\end{equation}
where \( \hat{p}_N \) and \( p_N \) are the predicted and true final positions.

Average Mahalanobis Distance (AMD) \cite{mohamed_social-implicit_2022} compares predicted samples to ground truth, accounting for the shape of the predicted distribution:
\begin{equation}
    \text{AMD} = \frac{1}{M N} \sum_{i=1}^{M} \sum_{t=1}^{N} \sqrt{(\hat{p}^i_t - p^i_t)^\top \Sigma^{-1} (\hat{p}^i_t - p^i_t)}
\end{equation}
where \( \hat{p}^i_t \) is the \( i \)-th predicted sample at time \( t \), \( p^i_t \) is the corresponding ground truth, \( \Sigma \) is the covariance matrix of the predictions, \( M \) is the number of samples, and \( N \) is the number of time steps.

Average Maximum Eigenvalue (AMV) \cite{mohamed_social-implicit_2022} quantifies prediction uncertainty by measuring spread in the predicted samples:
\begin{equation}
    \text{AMV} = \frac{1}{M N} \sum_{i=1}^{M} \sum_{t=1}^{N} \lambda_{\max}(\Sigma^i_t)
\end{equation}
where \( \lambda_{\max}(\Sigma^i_t) \) is the largest eigenvalue of the covariance matrix \( \Sigma^i_t \) for sample \( i \) at time \( t \).

\subsection{Robot Model}

The robot's motion is modeled using a non-holonomic kinematic unicycle model, which is commonly used in practice for robots with differential drive locomotion. The state of the robot is defined as $\mathbf{x} = [x, y, \theta]^\top$, where $(x, y)$ is the position and $\theta$ is the heading angle. The control inputs are linear and angular velocities $\mathbf{u} = [v, \omega]^\top$. The robot dynamics is:
\begin{subequations}\label{eq:kinematics}
\begin{align}
\dot{x} &= v \cos(\theta), \label{eq:kinematics_a} \\
\dot{y} &= v \sin(\theta), \label{eq:kinematics_b} \\
\dot{\theta} &= \omega, \label{eq:kinematics_c}
\end{align}
\end{subequations}
with control limits $v \in [-1.0, 1.0]$ m/s and $\omega \in [-1.0, 1.0]$ rad/s.

\section{Methodology} \label{sec:si_mpc}

This section describes how we combine the SI predictive model with an MPC local planner for mobile robot navigation in crowds.

\subsection{Social-Implicit: A Lightweight and Distribution-Aware Pedestrian Motion Prediction Model}

The \textit{Social-Implicit} model \cite{mohamed_social-implicit_2022} presents a state-of-the-art approach to pedestrian trajectory prediction, addressing common limitations in evaluation methods while offering a compact and efficient architecture suitable for real-time applications. Traditional metrics such as Best-of-N (BoN), ADE and FDE primarily focus on the single best trajectory prediction, often ignoring the distributional characteristics of the output. On the other hand, the SI model is trained using Implicit Maximum Likelihood Estimation (IMLE), which encourages the predicted trajectory samples to cluster around the ground truth distribution tightly. This results in improved prediction accuracy and reliability, which are very important characteristics for downstream modules such as the MPC local planner.

The SI model was chosen as a motion prediction module because it achieves a strong balance of prediction accuracy (lowest ADE/FDE), computational efficiency (5.8K parameters, sub-10ms inference), and ease of integration. SI’s simple architecture enables direct, parallelizable use in MPC frameworks without preprocessing or social pooling, while maintaining robust performance even in dense crowds. Also, smaller models are less prone to overfitting \cite{overfitting}. The model used for this paper is trained on the ETH dataset \cite{5459260}.

In the following, we describe how the SI model's output is integrated with an MPC, which is necessary for the setup. For each pedestrian $i = 1, \dots, M$, the observed history $h^i$ is a sequence of $N_h$ past positions
\[
h^i = \left( (x^i_{t-N_h+1}, y^i_{t-N_h+1}), \dots, (x^i_t, y^i_t) \right)
\]
This means that for pedestrian $i$, their 2D positions $(x, y)$ at each timestep from $t-N_h+1$ up to the current time $t$ exist\footnote{If the full history of a pedestrian is not observed at the current time, the first observed position is used to pad the history sequence to the necessary length.}. Each $h^i$ is a matrix in $\mathbb{R}^{N_h \times 2}$, where each row is a $(x, y)$ position at a past timestep. All pedestrian histories are stacked together into a single tensor 
$H = [h^1, \dots, h^M] \in \mathbb{R}^{M \times N_h \times 2}$.

The model predicts, for each pedestrian $i$, a future trajectory of length $N$:
\[
\hat{p}^i = \left( (\hat{x}^i_{t+1}, \hat{y}^i_{t+1}), \dots, (\hat{x}^i_{t+N}, \hat{y}^i_{t+N}) \right) \in \mathbb{R}^{N \times 2}
\]
This is the sequence of predicted future 2D positions for pedestrian $i$.

All predicted futures are stacked as:
\[
P = [\hat{p}^1, \dots, \hat{p}^M] \in \mathbb{R}^{M \times N \times 2},
\]
where $P$ is a tensor containing the predicted future trajectories for all pedestrians.  We denote
\[
P \sim \widetilde{p}_{\Theta}(\cdot \mid H)
\]
to indicate that \( \widetilde{p}_{\Theta} \) is our predictive model, parameterized by \( \Theta \), which defines a probability distribution over possible future trajectories \( P \) conditioned on the observed history \( H \).

\subsection{Model Predictive Control} \label{subsec:nominal_mpc}

The objective is to compute a collision-free and dynamically feasible trajectory that minimizes a cost function over a finite prediction horizon while accounting for the predicted motion of surrounding pedestrians. Let the robot's state at time $t$ be denoted by $\mathbf{x}_t \in \mathbb{R}^n$, and control inputs $\mathbf{u}_t \in \mathbb{R}^m$. The robot aims to find a trajectory $\{\mathbf{x}_t\}_{t=0}^N$ by solving an optimization problem of the form:
\begin{equation}
\min_{\mathbf{u}_{0:N-1}} \sum_{t=0}^{N-1} \ell(\mathbf{x}_t, \mathbf{u}_t) + \ell_N(\mathbf{x}_N)
\end{equation}
subject to:
\begin{align*}
\mathbf{x}_{t+1} &= f(\mathbf{x}_t, \mathbf{u}_t), \\
\mathbf{x}_t &\in \mathcal{X}_{\text{free}}(t), \\
\mathbf{u}_t &\in \mathcal{U}, \\
\forall t &\in \{0, \ldots,N-1\}, 
\end{align*}
where $\ell(\cdot)$ and $\ell_N(\cdot)$ denote the stage and terminal costs, $f(\cdot)$ is the discretized system dynamics, $\mathcal{U}$ is the set of admissible controls, and $\mathcal{X}_{\text{free}}(t)$ represents the obstacle-free space, which is time-varying due to dynamic agents such as pedestrians. 

We define the stage cost $\ell(\mathbf{x}_t, \mathbf{u}_t)$ as the following weighted quadratic form:
\begin{equation}
\ell(\mathbf{x}_t, \mathbf{u}_t) = 
\left\|
\mathbf{x}_t - \mathbf{x}_\text{goal}
\right\|^2_{Q}
+ \| \mathbf{u}_t \|^2_R
\end{equation}
where $\mathbf{x}_\text{goal}$ is the goal state, $Q$ and $R$ are weight matrices and $||\cdot||$ is the L2 norm. The terminal cost $\ell_N(\mathbf{x}_N) = \left\| \mathbf{x}_N - \mathbf{x}_\text{goal} \right\|^2_{Q_T}$ emphasizes reaching the goal, where $Q_T$ is a terminal weight matrix.

The predicted pedestrian trajectories are used to define $\mathcal{X}_{\text{free}}(t)$. To ensure collision avoidance, we impose circular safety zones of radii $r_r$ (robot) and $r_p$ (pedestrian) in the following way:

\begin{equation}
\left\| \left[ x_t, y_t \right] - \hat{p}^i_t \right\| \geq r_r + r_p, \, \forall i \in [1, M], \; \forall t \in [0,  N].
\end{equation}

\section{Performance Evaluation} \label{sec:results}

This section describes what hardware is used followed by the test setup and analysis. It presents the evaluation of the combined SI predictive model with an MPC local planner for mobile robot navigation.

\subsection{Hardware Setup}
The Continental's delivery robot is equipped with a comprehensive sensor suite and high-performance computing hardware to support real-time PTP and MPC. Key perception sensors include a 3D LiDAR Robosense RS-Helios 5515 for high-resolution environmental mapping and two Intel RealSense D435 depth cameras for RGB-D data acquisition. Computation is handled by an Intel NUC11PHKi7C Mini PC with an Intel Core i7 processor, 16GB RAM, and an NVIDIA RTX2060 GPU, enabling the execution of complex PTP models and MPC algorithms.

The MPC architecture is implemented using the CasADi optimization framework and ROS for system integration. The control loop operates at a 100 ms sampling time, using a 4th-order Runge-Kutta discretization of the robot's dynamics. Optimization is performed with IPOPT NLP solver, and CasADi’s code generation is used to compile efficient solver code for real-time execution. This setup ensures responsive control while managing the computational demands of integrating PTP models into dynamic navigation tasks.

\subsection{Test Environment and Layout}

\begin{figure}
    \centering
    \includegraphics[width=1\linewidth]{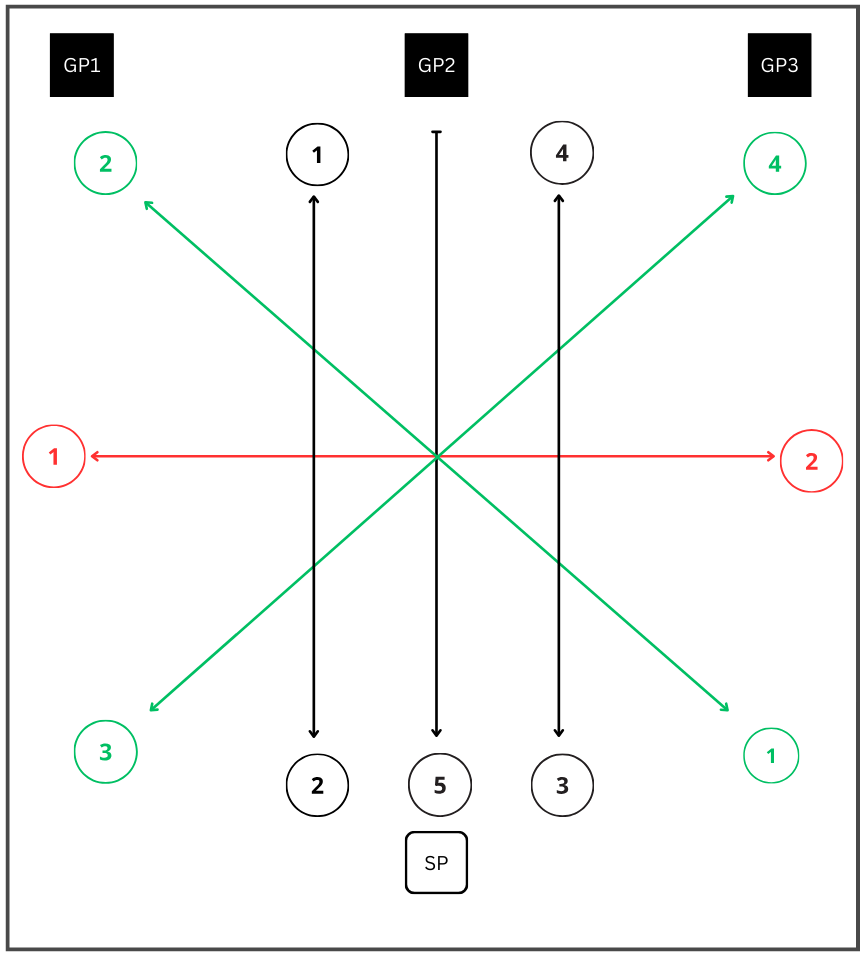}
    \caption{Schematic of the test environment showing robot paths (SP to GP) and pedestrian routes (color-coded).}
    \label{fig:1 Test Plan}
\end{figure}

To evaluate the navigation performance in crowds of the MPC planner with the learning-based predictive model, we set up different scenarios that represent real situations encountered during deployment. The robot begins at a fixed central start position (SP in Fig.~\ref{fig:1 Test Plan}) in all test cases. From SP, it navigates toward one of three goal positions (GP1, GP2, GP3), representing distinct real-world destinations as shown in Fig.~\ref{fig:1 Test Plan}:
\begin{itemize}
    \item GP1 (top-left): Simulates a leftward navigation.
    \item GP2 (top-center): Represents a straightforward corridor-like route.
    \item GP3 (top-right): Simulates a rightward navigation.
\end{itemize}

\begin{figure}
    \centering
    \includegraphics[width=1\linewidth]{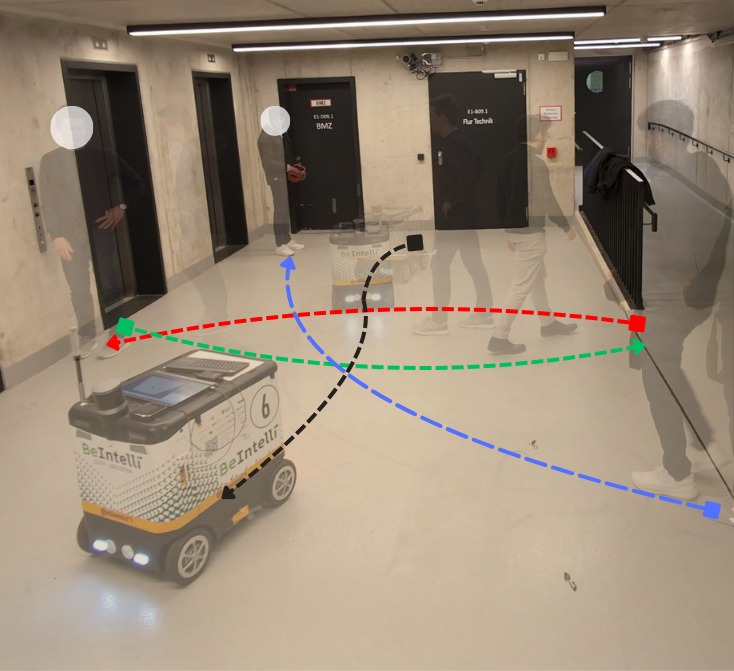}
    \caption{Visualization of the robot navigation with the SI-MPC planner from the 8th experiment. The black arrow represents the robot’s motion, while the motion of pedestrians is illustrated by the red, green, and blue arrows.}
    \label{fig:Path Taken by the robot and pedestrians in real time}
\end{figure}

Pedestrians follow predefined routes to introduce structured interaction scenarios. Their paths are categorized by movement direction (see Fig.~\ref{fig:1 Test Plan}):
\begin{itemize}
    \item Red paths: Horizontal (left-right or right-left).
    \item Green paths: Diagonal.
    \item Black paths: Vertical (toward or away from the robot).
\end{itemize}
Fig.~\ref {fig:Path Taken by the robot and pedestrians in real time} visualizes one particular scenario from the set of conducted experiments, while a more detailed description of the experiments is provided in Table~\ref{tab:robot_pedestrian_tests}.

\begin{table*}[tbh]
    \centering
    \caption{Description of the conducted experiments.}
    \label{tab:robot_pedestrian_tests}
    \scriptsize
    \begin{tabularx}{\textwidth}{|c|X|X|X|X|}
        \hline
        \textbf{Test Case} & \textbf{Robot Path} & \textbf{Pedestrian 1} & \textbf{Pedestrian 2} & \textbf{Pedestrian 3} \\
        \hline
        Scene 1 & SP → GP1 & \textcolor{red}{Red 1} → \textcolor{red}{Red 2} & — & — \\
        Scene 2 & SP → GP2 & GP2 → \textcolor{black}{Black 5} & — & — \\
        Scene 3 & SP → GP3 & \textcolor{green!50!black}{Green 2} → \textcolor{green!50!black}{Green 1} & — & — \\
        Scene 4 & SP → GP1 & \textcolor{red}{Red 2} → \textcolor{red}{Red 1} & \textcolor{black}{Black 1} → \textcolor{black}{Black 2} & — \\
        Scene 5 & SP → GP2 & \textcolor{red}{Red 1} → \textcolor{red}{Red 2} & \textcolor{green!50!black}{Green 4} → \textcolor{green!50!black}{Green 3} & — \\
        Scene 6 & SP → GP3 & \textcolor{black}{Black 4} → \textcolor{black}{Black 3} & \textcolor{green!50!black}{Green 3} → \textcolor{green!50!black}{Green 4} & — \\
        Scene 7 & SP → GP1 & \textcolor{black}{Black 1} → \textcolor{black}{Black 2} & GP2 → \textcolor{black}{Black 5} & \textcolor{green!50!black}{Green 4} → \textcolor{green!50!black}{Green 3} \\
        Scene 8 & SP → GP2 & \textcolor{red}{Red 2} → \textcolor{red}{Red 1} & \textcolor{red}{Red 1} → \textcolor{red}{Red 2} & \textcolor{black}{Black 1} → \textcolor{black}{Black 2} \\
        Scene 9 & SP → GP1 & \textcolor{black}{Black 1} → \textcolor{black}{Black 2} & \textcolor{green!50!black}{Green 2} → \textcolor{black}{Black 1} & \textcolor{red}{Red 1} → \textcolor{red}{Red 2} \\
        Scene 10 & SP → GP2 & \textcolor{black}{Black 4} → \textcolor{black}{Black 3} & \textcolor{green!50!black}{Green 1} → \textcolor{black}{Black 2} & GP2 → \textcolor{black}{Black 5} \\
        \hline
    \end{tabularx}
\end{table*}

Test cases systematically varied the robot’s goal position and pedestrian density. Pedestrians crossed paths with the robot, moved in parallel, or approached from the opposite direction. Also, the pedestrians always tend to reach their goal positions that are specified at the current test scenario. The tests aim to evaluate the robot’s navigation performance under different interaction scenarios and pedestrian densities (1–3 pedestrians). The experiments consist of 20 test cases, divided into two categories -- 10 experiments using the SI model and 10 experiments with the CV model.

We analyze several aspects of navigation performance with this setup, including the comparison of robot behavior under SI and CV models using metrics such as ADE, FDE, AMD, and AMV. Also, we assess how the choice of predictive model influences time-to-goal and minimum distance to pedestrians. This controlled test environment ensures a fair comparison of navigation strategies and highlights the robot’s capacity to adapt to varying pedestrian interactions.

\subsection{Quantitative Analysis}

The performance evaluation of the SI and CV models integrated in MPC for motion planning of a robot provides critical insights into their effectiveness in dynamic pedestrian environments. This part of the analysis focuses on quantitative metrics (prediction accuracy, motion planning efficiency, safety margins) and qualitative observations (smoothness, adaptability, real-world applicability). The results highlight key trade-offs between safety, efficiency, and motion stability, offering actionable recommendations for deploying these models in real-world robotics.

\begin{table}[tbh]
    \centering
    \small
    \caption{Trajectory prediction metrics for varying numbers of pedestrians.}
    \resizebox{\columnwidth}{!}{%
        \begin{tabular}{llccc}
            \toprule
            \textbf{\# Ped.} & \textbf{Metric} & \textbf{SI} & \textbf{CV} & \textbf{Improv. (\%)} \\
            \midrule

            \multirow{4}{*}{1}
            & ADE  & \textbf{0.455} & 1.655 & 72.5 \\
            & FDE  & \textbf{0.687} & 2.849 & 75.9 \\
            & AMD  & \textbf{4.430} & 8.151 & 45.6 \\
            & AMV  & \textbf{0.018} & 0.122 & 84.8 \\
            \midrule

            \multirow{4}{*}{2}
            & ADE  & \textbf{0.561} & 0.746 & 24.7 \\
            & FDE  & \textbf{0.859} & 1.338 & 35.8 \\
            & AMD  & 5.367 & \textbf{3.226} & -66.4 \\
            & AMV  & \textbf{0.018} & 0.120 & 84.5 \\
            \midrule

            \multirow{4}{*}{3}
            & ADE  & \textbf{0.560} & 0.938 & 40.3 \\
            & FDE  & \textbf{0.873} & 1.654 & 47.2 \\
            & AMD  & 5.627 & \textbf{4.443} & -26.7 \\
            & AMV  & \textbf{0.018} & 0.121 & 84.6 \\
            \bottomrule
        \end{tabular}%
    }
    \label{tab:ptp_metrics_table}
\end{table}

In low-density environments, the SI model shows a clear advantage over the CV model, as shown in Table~\ref{tab:ptp_metrics_table}. It reduces ADE (0.455 vs. 1.655) and FDE (0.687 vs. 2.849) by 72.5\% and 75.9\%, respectively, reflecting significantly more accurate trajectory predictions. This stems from the SI model’s ability to anticipate pedestrian intent through social navigation behaviors, rather than assuming constant motion. Additionally, the SI model yields an 84.8\% lower AMV (0.018 vs. 0.122), indicating greater confidence and reduced uncertainty, and a 45.6\% lower AMD (4.430 vs. 8.151), showing closer alignment with true pedestrian distributions. These results highlight the SI model’s superior reliability in simple, predictable settings.

As interaction complexity increases, the SI model maintains better results in ADE (0.561 vs. 0.746, 24.7\% improvement) and FDE (0.859 vs. 1.338, 35.8\% improvement), though the margin narrows compared to the 1-pedestrian case. This is expected, as multi-pedestrian interactions introduce more variables such as crossing paths and group dynamics. 
However, the AMD (5.367 vs. 3.226) increases by 66.4\% for the SI model, indicating a broader, more conservative prediction space. This suggests the SI model accounts for a wider range of possible pedestrian behaviors to avoid collisions, even if it slightly overestimates uncertainty. The AMV remains 84.5\% lower (0.018 vs. 0.120), reinforcing that the SI model’s predictions are still more stable and less erratic than the CV model’s.

In high-density environments, the SI model continues to outperform the CV model in ADE (0.560 vs. 0.938, 40.3\% improvement) and FDE (0.873 vs. 1.654, 47.2\% improvement). These reductions in error demonstrate the SI model’s robustness in highly dynamic settings, where pedestrians may abruptly change direction or speed. Like the 2-pedestrian case, the AMD (5.627 vs. 4.443) is 26.7\% higher for the SI model, further emphasizing its cautious approach in crowded scenarios. This conservative behavior ensures safer navigation but may slightly reduce path optimality. The AMV (0.018 vs. 0.121) remains 84.6\% lower, confirming that the SI model’s predictions are statistically more reliable despite the increased complexity.

\begin{table}[!tbh]
    \centering
    \caption{Comparison of ADE, FDE, AMD, and AMV for open- and closed-loop evaluation.}
    \label{tab:eth_vs_si}
    \resizebox{\columnwidth}{!}{%
        \begin{tabular}{lcccc}
            \toprule
            \textbf{Method} & \textbf{ADE} & \textbf{FDE} & \textbf{AMD} & \textbf{AMV} \\
            \midrule
            Open Loop (ETH data)     & 0.642 & 1.379 & 3.047 & 0.127 \\
            Closed Loop (SI-MPC)    & 0.525 & 0.806 & 5.141 & 0.018 \\
            \bottomrule
        \end{tabular}%
    }
\end{table}

Table~\ref{tab:eth_vs_si} reveals a nuanced relationship between open-loop benchmark performance and real-world behavior in a full navigation stack. The SI model, trained solely on the ETH dataset (early stopping at epoch 45, minimum validation loss 0.053), achieves better trajectory accuracy (ADE: 0.525 vs. 0.642; FDE: 0.806 vs. 1.379) but shows higher AMD (5.141 vs. 3.047), indicating a broader predictive distribution during operation. Its much lower AMV (0.018 vs. 0.127) further suggests a more cautious, stable behavior in real-time planning. This highlights a key insight: strong open-loop metrics do not guarantee equivalent closed-loop performance. Integration into a navigation stack introduces constraints—like real-time decision-making and safety—that alter model behavior in ways not captured by static evaluations. Thus, open-loop metrics, while useful, must be paired with system-level evaluations to assess true operational generalization.

\begin{table}[!tbh]
    \centering
    \small
    \caption{Closed-loop performance comparison of SI and CV motion prediction methods.}
    \resizebox{\columnwidth}{!}{%
        \begin{tabular}{llccc}
            \toprule
            \textbf{\# Ped.} & \textbf{Metric} & \textbf{SI} & \textbf{CV} & \textbf{Improv. (\%)} \\
            \midrule

            \multirow{4}{*}{1} 
            & Min. Dist. (m)       & \textbf{0.50}  & 0.31  & 61.80 \\
            & Time Taken (s)       & \textbf{14.80} & 18.77 & 21.10 \\
            & Jerk (m/s$^3$)       & \textbf{0.13}  & 0.19  & 32.35 \\
            & MPC MSE              & 0.02           & 0.02  & 12.54 \\
            \midrule

            \multirow{4}{*}{2} 
            & Min. Dist. (m)       & \textbf{0.30}  & 0.22  & 36.70 \\
            & Time Taken (s)       & 24.82          & \textbf{19.76} & -25.50 \\
            & Jerk (m/s$^3$)       & 0.17           & 0.17  & 1.08  \\
            & MPC MSE              & 0.024          & \textbf{0.020} & -23.08 \\
            \midrule

            \multirow{4}{*}{3} 
            & Min. Dist. (m)       & \textbf{0.26}  & 0.19  & 41.10 \\
            & Time Taken (s)       & 21.22          & \textbf{18.65} & -13.80 \\
            & Jerk (m/s$^3$)       & \textbf{0.21}  & 0.30  & 28.38 \\
            & MPC MSE              & \textbf{0.03}  & 0.04  & 18.61 \\
            \bottomrule
        \end{tabular}%
    }
    \label{tab:robot_performance_metrics}
\end{table}

The comparative analysis in Table~\ref{tab:robot_performance_metrics} of minimum distance and time taken reveals key behavioral differences between the SI and CV models. In single-pedestrian scenarios, the SI model maintained a 61.8\% larger safety margin (0.50\,m vs. 0.31\,m) while completing the trajectory 21.1\% faster (14.80\,s vs. 18.77\,s), due to its proactive adaptation to pedestrian movement. As pedestrian density increased, the SI model prioritized safety over speed: in two-pedestrian cases, it preserved a 36.7\% larger distance (0.30\,m vs. 0.22\,m) but took 25.5\% more time (24.82\,s vs. 19.76\,s); in three-pedestrian scenarios, it achieved a 41.1\% safety gain (0.268\,m vs. 0.190\,m) with a 13.8\% time increase (21.22\,s vs. 18.65\,s). These results highlight the practical advantages of an MPC planner using the SI model.

Motion smoothness, measured by jerk (rate of acceleration change), directly influences pedestrian trust. In single-pedestrian tests, the SI model reduced jerk by 31.6\% (0.13\,m/s$^3$ vs. 0.19\,m/s$^3$), indicating smoother velocity transitions. For two pedestrians, both models showed similar jerk (0.17\,m/s$^3$), though the SI model had more consistent acceleration patterns (e.g., gradual deceleration in Test~6 vs. abrupt stops in CV’s Test~4). The SI model’s advantage was more evident in dense scenarios: with three pedestrians, it lowered jerk by 28.4\% (0.21\,m/s$^3$ vs. 0.30\,m/s$^3$), avoiding CV’s erratic behavior (e.g., sharp deceleration in Test~7). This aligns better with human locomotion norms, reducing discomfort and improving social acceptance.

Prediction reliability, measured via Mean Squared Error (MSE), further distinguishes the models. The SI model outperformed CV in 8 of 10 tests, with a 66.7\% lower MSE in Test 1 (0.005 vs. 0.017). This accuracy stems from its socially aware trajectory forecasting, which accounts for pedestrian intent. Notably, the CV model excelled in Test 10 (MSE: 0.026 vs. SI’s 0.049), likely due to the SI model’s overcompensation in dense crowds. While this outlier suggests occasional over-cautiousness, the SI model’s overall lower MSE (e.g., 37.6\% improvement in Test 5) confirms its robustness in most real-world conditions.

\subsection{Qualitative Analysis}

\begin{figure}
    \centering
    \includegraphics[width=1\linewidth]{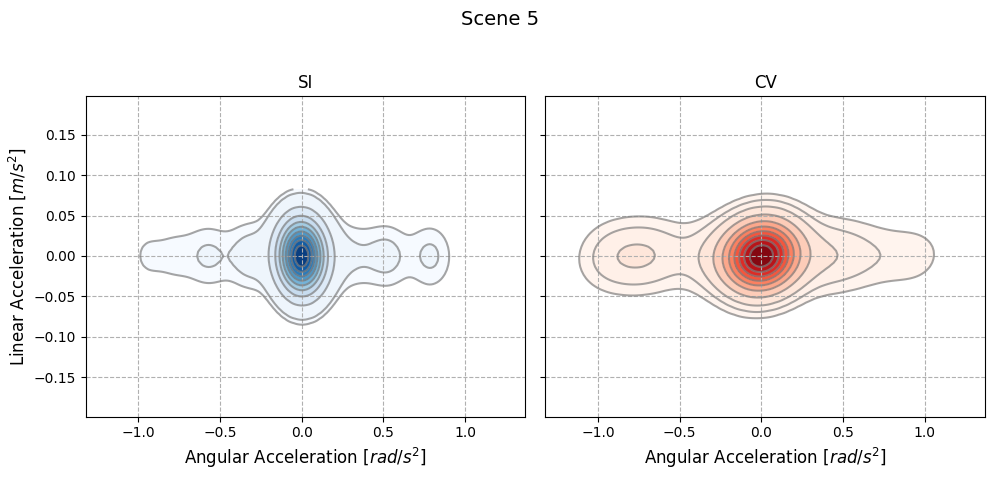}
    \includegraphics[width=1\linewidth]{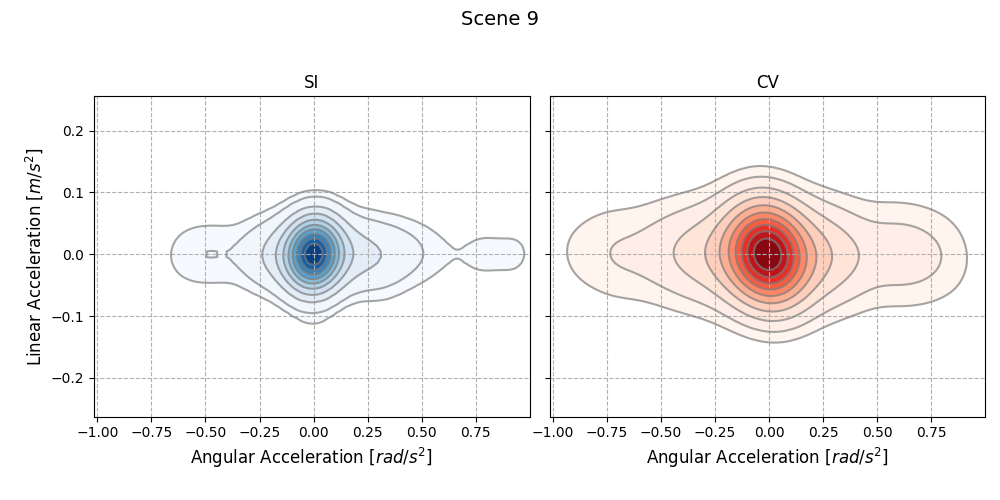}
    \caption{(Top) Linear vs. angular acceleration contour plots for Scene 5. (Bottom) Linear vs. angular acceleration contour plots for Scene 9.}
    \label{fig:Acceleration Plots}
\end{figure}

Fig.~\ref{fig:Acceleration Plots} shows joint linear versus angular acceleration contour plots for two pedestrian interaction scenes. In both, the SI-based MPC produces more compact, elliptical contours centered near the origin, indicating lower control effort and more stable acceleration. This suggests that socially informed predictions help the robot anticipate future states better, reducing the need for corrective maneuvers. In contrast, the CV-based MPC yields broader, more asymmetric distributions, especially in angular acceleration, reflecting more variable and reactive control inputs due to its limited modeling of pedestrian behavior. This effect is pronounced in Scene 9, where higher angular acceleration variance indicates frequent adjustments and poor predictive accuracy. These findings imply that socially aware prediction improves trajectory smoothness and control efficiency, while naive models like CV increase instability and control burden, potentially compromising safety and comfort in close human-robot interactions.

\begin{figure*}
    \centering
    \includegraphics[trim = 3mm 0mm 4mm 0mm, clip, width=1\linewidth]{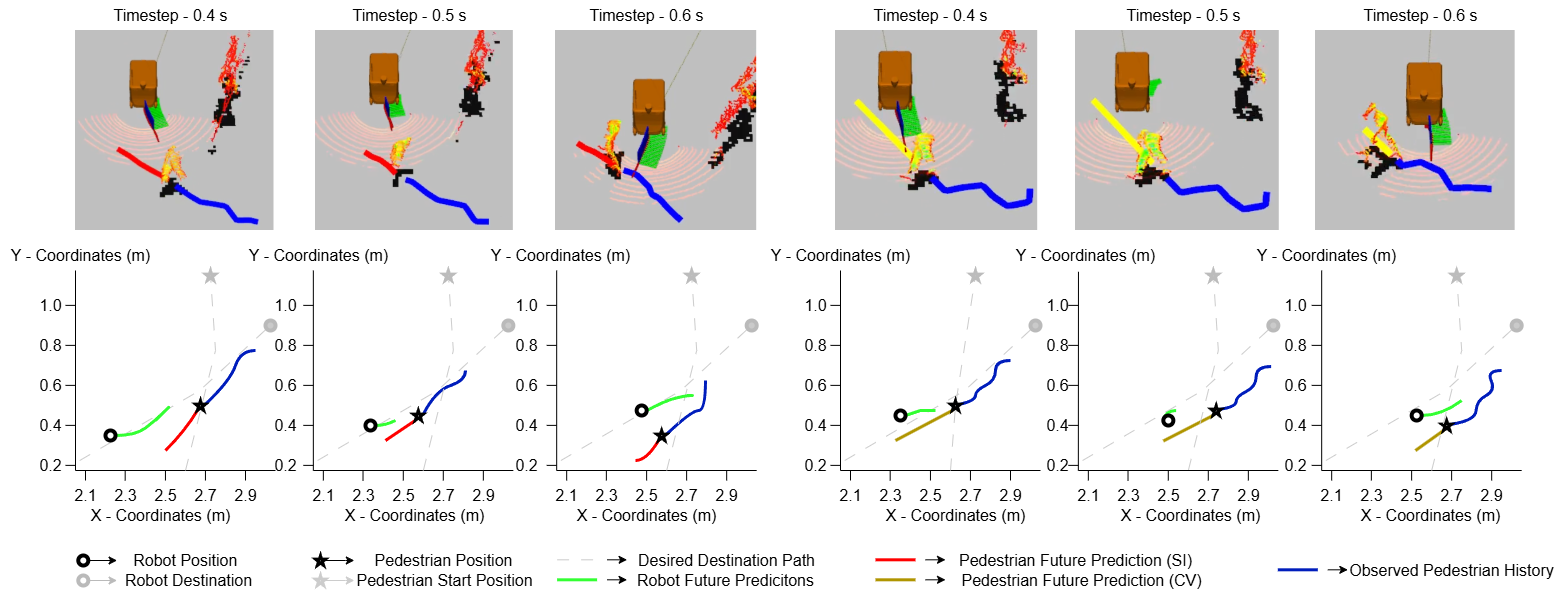}
    \caption{Robot navigation (Scene 1) through crowded environments using SI and CV models. Top row: Runtime snapshots from RVIZ.  Bottom row: Corresponding top-down schematic representations of trajectories.}
    \label{fig: RVIZ and Plot Representation of comparison of CV and SI with Robot Navigation}
\end{figure*}

Figure~\ref{fig: RVIZ and Plot Representation of comparison of CV and SI with Robot Navigation} illustrates Scene 1, where the robot navigates from SP to GP1 as a pedestrian moves from Red 1 to Red 2 under both CV and SI prediction models (see Table~\ref{tab:robot_pedestrian_tests} and Fig.~\ref{fig:1 Test Plan}). The visualization shows that the CV model, due to its simplistic and localized predictions, prompts the robot to halt abruptly when a potential collision is detected, particularly at 0.5s and 0.6s. In contrast, the SI model enables smoother, adaptive behavior: the robot reduces speed and adjusts its path gradually, reflecting a more temporally consistent understanding of pedestrian motion. This qualitative difference aligns with quantitative metrics, especially jerk values, where lower values in the SI case confirm its advantage.

Additionally, the overall time taken to reach the goal with SI-MPC is more reasonable and efficient, as it avoids unnecessary stops and abrupt actions, thereby aligning with the observed improvements in time-to-goal metrics presented in the quantitative analysis. Thus, this scene effectively highlights the advantages of the SI model in generating socially compliant and efficient navigation for the robot.

\section{Conclusion}\label{sec:conclusion}
This study presents three key advances in robot navigation using a deep learning-based Social-Implicit (SI) predictor integrated with MPC on a real robot. First, the closed-loop system demonstrates real-time feasibility at 10~Hz, validating stable MPC performance with neural predictors. Second, comparisons reveal a mismatch between open-loop and closed-loop behavior---despite improved prediction accuracy (e.g., ADE: 0.525 vs. 0.642), closed-loop deployment led to broader, more cautious predictions, emphasizing the need for system-level evaluation. Third, across varied pedestrian densities, the SI-MPC outperformed the CV-MPC approach, improving safety margins (by 41--75\%), reducing motion jerk (28--84\%), and achieving faster goal attainment (21--47\%), with only moderate efficiency trade-offs in dense crowds. Overall, learned predictors enhance real-world safety and social compliance, though future work should address scalability and conservatism-efficiency trade-offs.

\section*{\uppercase{Acknowledgements}}

This work was funded by the German Federal Ministry for Economic Affairs and Climate Action within the project \textit{nxtAIM}.


\bibliographystyle{apalike}
{\small
\bibliography{main.bbl}}

\end{document}